# A Novel Approach for Image Segmentation based on Histograms computed from Hue-data

**Viraj Mavani**

Electronics and Communication
Department, L. D. College of Engineering
*viraj.mavani1996@gmail.com*

**Ayesha Gurnani**

Electronics and Communication
Department, L. D. College of Engineering
*ayeshagurnani1302@gmail.com*

**Jhanvi Shah**

Electronics and Communication
Department, L. D. College of Engineering
*jhanvishah59@gmail.com*

*Abstract*–Computer Vision is growing day by day in terms of user specific applications. The first step of any such application is segmenting an image. In this paper, we propose a novel and grass-root level image segmentation algorithm for cases in which the background has uniform color distribution. This algorithm can be used for images of flowers, birds, insects and many more where such background conditions occur. By image segmentation, the visualization of a computer increases manifolds and it can even attain near-human accuracy during classification.

## I. INTRODUCTION

Image Segmentation is the first and an essential part of any computer vision problem. Due to the possibility of Computer Vision being applied to various day to day problems faced by humans, new methods to tackle them have become important. Image segmentation basically means extracting key information from an image for further processing while discarding the rest.

At the very beginning, we have proposed a new algorithm for image segmentation based on color details of an image. Our algorithm can be applied efficiently to those cases where major contributing feature of the subject is color and when there exists some amount of contrast between foreground and background. This algorithm follows an iterative approach for background removal thereby highlighting the subject.

We have performed experiments and have used our image segmentation algorithm on the Oxford University 102 Flower Category Database[1] and then further applied it to images from the Visipedia CUB 200-2011[2] Dataset. We got fine results in which the background of the image was completely removed and key foreground data was extracted.

## II. PROPOSED METHOD

In our proposed method, the underlying assumption is that the background of the naturally occurring scene has uniformity in hue-value. A step wise description of removing background from the image to get the subject of interest is given below:

1. We first crop the boundaries of the image matrix having pre-defined width and height.

2. The 3 channel RGB images of boundaries obtained after cropping are then converted into the HSI Color Space Representation.

3. Now histograms with 256 bins of the 2-dimensional hue channel are evaluated for each edge of the image.

4. We then compute the index of those bins in the histograms which have number of pixels above some predefined threshold. By doing this, we end up with the hue-values of those pixels which occur frequently in the boundary images.

5. By pixel wise iteration on the hue channel of the whole image, we can remove the pixels corresponding to the background obtained from the previous step.

At the end of this process, we are left with a segmented image with some amount of salt and pepper noise. This noise is nothing but those pixels of the background which are different from the majority. This happens because not all pixels of the background are close in terms of hue values. Plus, some data from the foreground might also have colors close to background resulting into a grainy image in majority of the cases. This noise can be removed by using a median filter[3]. This method can be applied to problems like flower category classification, bird species categorization or classification of insects because they are found in environmental conditions resembling the underlying assumption for the background in our algorithm. This algorithm presents clean segmentation when there is sufficient contrast between the foreground and the background.

III. APPLICATION IN IMAGES OF FLOWERS

In cases of flowers, the most distinguishing feature is color and so we have applied our algorithm for flower image segmentation. We used the Oxford University 102 Flower Category Database[1] and did experimental analysis to evaluate the performance of our algorithm. The pictorial representation of the process has been given in figure 2.

Here, the histogram has been plotted in step 2 of the method. This histogram corresponds to the hue-value of only one edge from the whole boundary of a 256x256 image. We can deduce from the histogram plot that a threshold value of 5 pixels per bin is suitable to determine frequent colors occurring in the background. These colors are removed by pixel-wise iteration thereby leaving us with only the essential foreground information.

We can see that the process gives a fine output when the background is uniform in context to color. We have given other examples, see Figure 1, where the algorithm does a good job in segmentation of flower images.

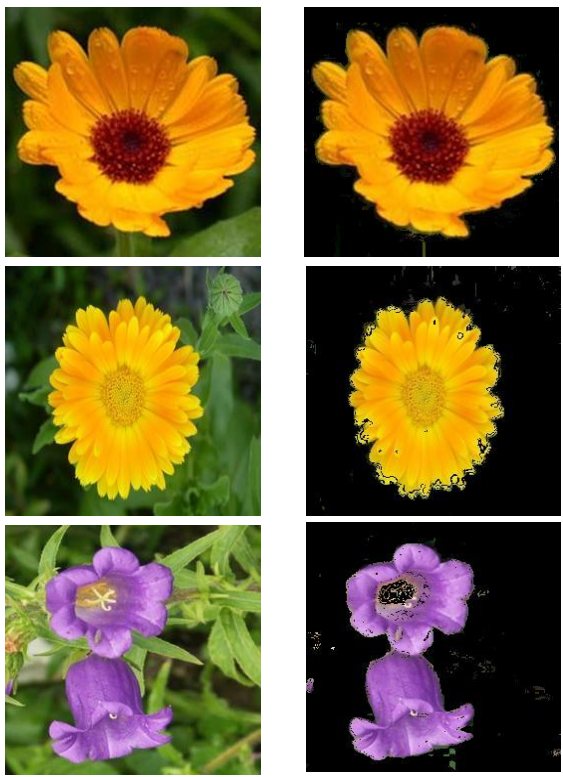

Figure 1: Example of image segmentation. (Left) Original Images (Right) Segmented Images

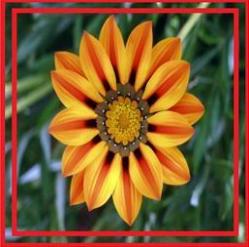

Step 1: Cropping the boundaries of the image.

Step 2: Converting the image from RGB Color Space to HSI Color Space Representation to evaluate the histogram

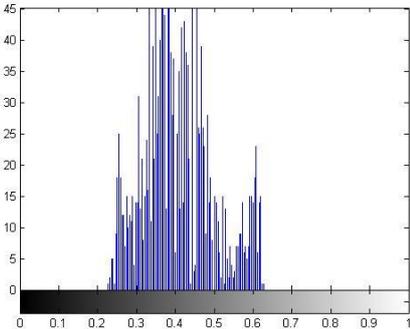

Step 3: Evaluating the histogram with 256 bins of the edges of the image. Here, the histogram of the hue value of a single edge has been evaluated and plotted for representaiton purpose.

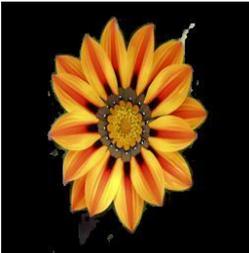

Step 4: The segmented image after iterative background pixel removal process.

Figure 2: The complete procedure for image segmentation by proposed algorithm.

## IV. COMPARISON WITH OTHER SEGMENTATION METHOD

We have compared the results of our method with the segmentation method used for the Oxford University 102 Flower Category Database[1] proposed by Maria-Elena Nilsback and Andrew Zisserman[4] which uses MRF cost function optimized using Graph-cuts. Our algorithm on the other hand has some limitations but gives accurate and required results in all applicable cases. The comparison has been showcased pictorially in figure 2. The results of Nilsback and Zisserman's method[4] are provided which are very good in some cases but in some they fail to differentiate between the background and foreground thus resulting into a completely blank image. On the other side, in some cases like the one shown in the second comparison example, our proposed algorithm fails to segment the image properly because of non-ideal background conditions while their method gives very fine result.

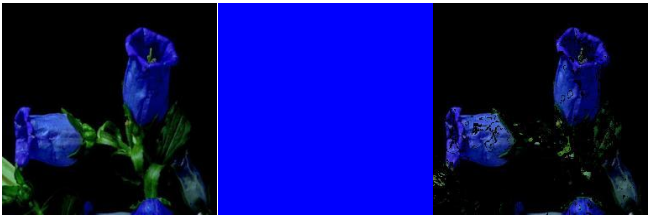

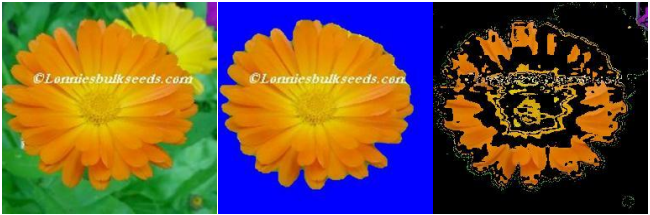

Figure 3: Comparison between Nilsback and Zisserman's Method[4] and the proposed method.

## V. OTHER POSSIBLE APPLICATIONS

In addition to flower image segmentation, the proposed method can be put to use for bird image segmentation, insect image segmentation and the like as these subjects also have color as their prominent feature. We applied the algorithm to images of birds from the Visipedia CUB 200-2011 Bird Species Dataset[2] which is curated by Pietro Perona's Vision Group at Caltech and Serge Belongie's Vision Group at Cornell Tech. By keeping the threshold values same as those applied to flower images, we got satisfactory results which are portrayed below. In the figure 4.1, due to uniform background and proper lighting conditions, the segmented image obtained is satisfactory whereas in figure 4.3, due to comparable hue values of background and the body of the bird, few pixels have been removed in the segmented image.

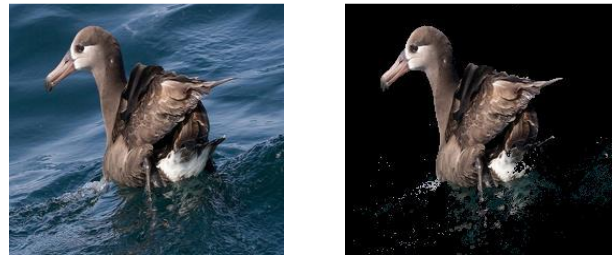

Figure 4.1

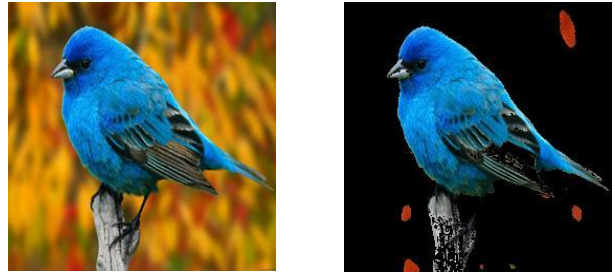

Figure 4.2

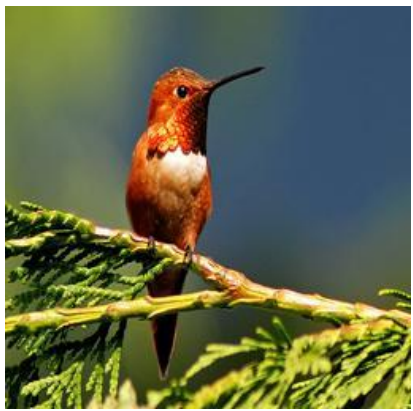 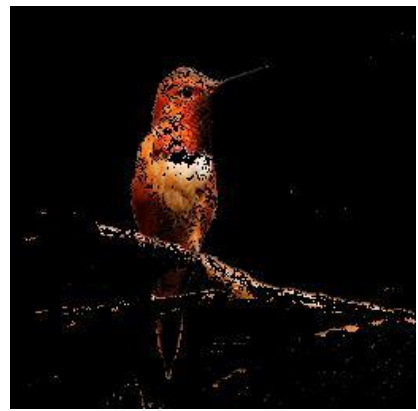

Figure 4.3: The pictures have been scaled for effective representation of the error.

## VI. CONCLUSION

In this paper a new method of segmenting an image is proposed. The given method removes the background thereby extracting the subject. We compared it with the results of MRF cost function optimized using Graph-cuts[4] and found that there are some cases where our algorithm works better than it while there are some cases where our algorithm has its own limitations when the background is non-uniform. Overall, the proposed method gives satisfactory results and can be used for further research work in the field of Computer Vision especially at a pre-classification stage of problems that are described in the paper.

## VII. ACKNOWLEDGMENTS